\documentclass[10pt,twocolumn,letterpaper]{article}

\usepackage[final]{wacv} 
\usepackage{graphicx}
\usepackage{amsmath}
\usepackage{amssymb}
\usepackage{booktabs}

\usepackage[pagebackref,breaklinks,colorlinks]{hyperref}

\usepackage[capitalize]{cleveref}
\usepackage[dvipsnames]{xcolor}
\usepackage{arydshln}
\usepackage{enumitem}

\def\model{\texttt{CusConcept}\xspace}

\crefname{section}{Sec.}{Secs.}
\Crefname{section}{Section}{Sections}
\Crefname{table}{Table}{Tables}
\crefname{table}{Tab.}{Tabs.}

\def\wacvPaperID{384} 
\def\confName{WACV}
\def\confYear{2025}

\newcommand{\hsz}[1]{\textcolor{black}{#1}}
\newcommand{\xz}[1]{\textcolor{black}{#1}}
\newcommand{\xzadd}[1]{\textcolor{black}{#1}}

\title{\xzadd{CusConcept: Customized Visual Concept Decomposition with Diffusion Models}}

\author{Zhi Xu\textsuperscript{1}\thanks{Equal contribution \quad $^\dagger$ Corresponding author}
\qquad
Shaozhe Hao$^{2*}$
\qquad
Kai Han$^{2\dagger}$
\vspace{0.3em} 
\\
{\normalsize \textsuperscript{1}Zhejiang University} \qquad  
{\normalsize \textsuperscript{2}The University of Hong Kong}
}

\begin{document}

\maketitle

\begin{abstract}
Enabling generative models to decompose visual concepts from a single image is a complex and challenging problem. In this paper, we study a new and challenging task, customized concept decomposition, wherein the objective is to leverage diffusion models to decompose a single image and generate visual concepts from various perspectives.
To address this challenge, we propose a two-stage framework, \model (short for \underline{Cus}tomized Visual \underline{Concept} Decomposition), to extract customized visual concept embedding \xzadd{vectors} that can be embedded into prompts for text-to-image generation.
In the first stage, \model employs a vocabulary-guided concept decomposition mechanism to build vocabularies along human-specified conceptual axes. The decomposed concepts are obtained by retrieving corresponding vocabularies and learning anchor weights.
In the second stage, joint concept refinement is performed to enhance the fidelity and quality of generated images. 
We further curate an evaluation benchmark for assessing the performance of the open-world concept decomposition task.
Our approach can effectively generate high-quality images of the decomposed concepts and produce related lexical predictions as secondary outcomes.
Extensive qualitative and quantitative experiments demonstrate the effectiveness of \model. Our code and data are available at \href{https://github.com/xzLcan/CusConcept}{https://github.com/xzLcan/CusConcept}.
\end{abstract}

\section{Introduction}
\label{sec:intro}

The compositionality and contextuality of concepts are important in human intelligence~\cite{sense1948gottlob,misra2017red}. 
In terms of compositionality, a complex concept, usually a physical entity, is the combination of multiple basic concepts; \xzadd{in} terms of contextuality, any basic concept cannot be created without context of basic concepts. Inpired by these properties, previous research~\cite{li2020symmetry, mancini2021open, mancini2022learning, misra2017red, naeem2021learning, nagarajan2018attributes, purushwalkam2019task, wei2019adversarial, li2022siamese, hao2023learning, ruis2021independent, saini2022disentangling, nayak2023learning, zheng2024caila, Lu_2023_CVPR} models a physical entity using the ``object'' concept, and the ``attribute'' \xzadd{concept} that describe the abstract object from different axes. Together, these two components form a complex and human-perceptible concept.

\begin{figure}[tb]
  \centering
  \includegraphics[width=0.99\linewidth]{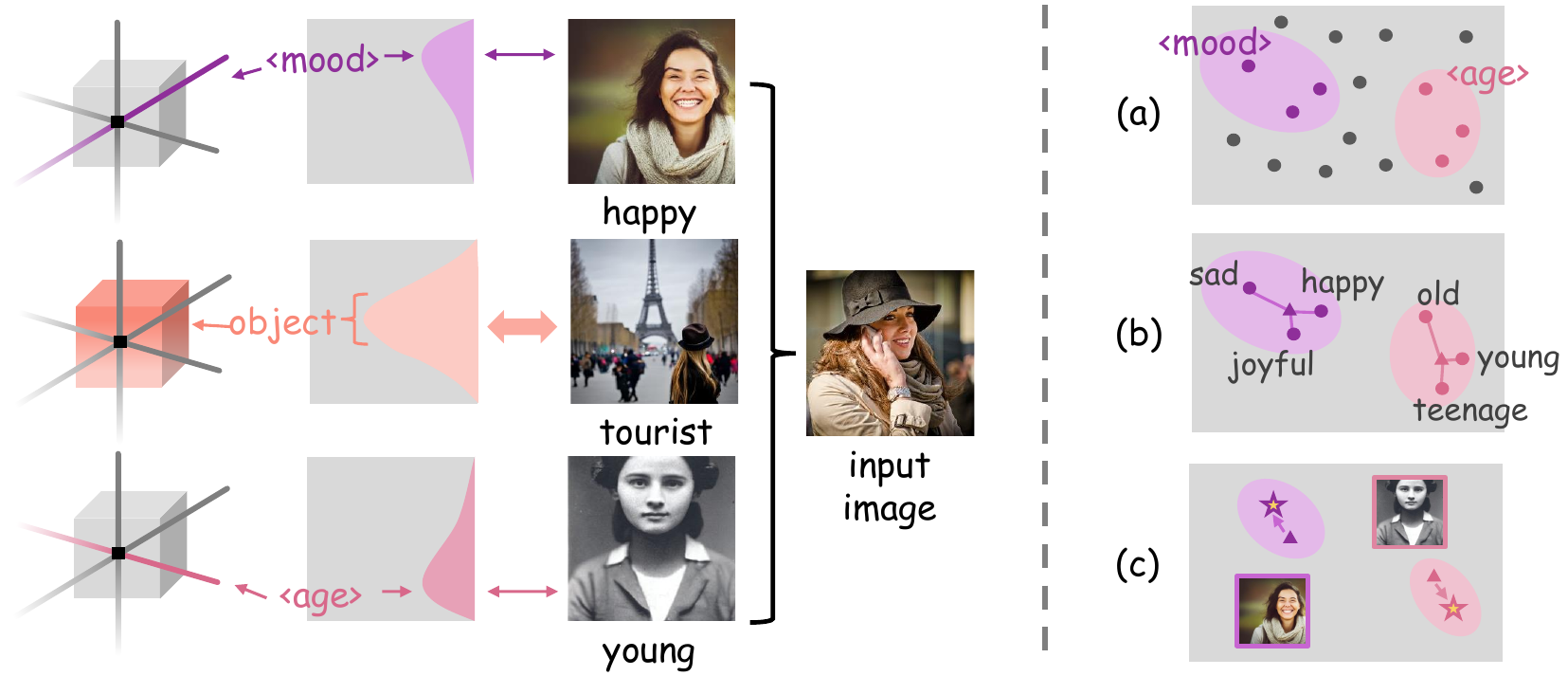}
  \caption{\textbf{Customized concept decomposition.} Our aim is to decompose the input image into the object concept and the attribute concepts along user-specified axes. \textbf{Left}: 
  We consider each visual entity to be the composition of the ``object'' concept and multiple ``attributes'' defined along different attribute axes. 
  Each disentangled concept, including the object and its attributes, has a domain, here simplified as one-dimensional probability distributions. \textbf{Right}: We illustrate the learning of concept embeddings in the 2D space. 
  \xzadd{
  (a) The words are distributed in the embedding space (gray dots), with words along the same attribute axis marked with the same color,~\eg, pink for age.
  (b) The word embeddings are combined using a weighted sum, similar to finding the centroids (triangles) of the same color dots in the space.
  (c) The weighted sum embeddings are further fine-tuned into final concept embeddings, like moving from triangles to stars in the space.
  }}
  \label{fig:intro}
\end{figure}
When humans define a physical entity, the ``object'' component is often apparent and certain. However, defining the ``attribute'' component from different axes poses significant variability. For example, a girl can be characterized as ``happy'' from the axis of mood, or ``young'' from the axis of age. Differentiating ``happy'' and ``\xzadd{young}'' conceptually is simple for humans, but quite challenging for \xzadd{machine learning models}.
In this paper, we \xzadd{disentangle} the concepts of ``object'' and axis-wise ``attributes'' from the perspective of \emph{generation}.

Recently, some works have addressed the related research topic of the disentanglement of compositional concepts with the use of generative diffusion models. 
\cite{chefer2023hidden} studies decompose images with word guidance.
\cite{vinker2023concept} explores different axes of a visual concept to learn multiple token embeddings.
\cite{lee2023languageinformed} disentangles different concept axes such as ``object'', ``color'', and ``material'' from a given image. 
These works highlight the potential for exploring various concept axes in the context of generative problems. However, 
they either lack the ability to steer the model towards learning the concept axes specified by humans~\cite{chefer2023hidden, vinker2023concept}, or they are constrained by a limited range of axes, thereby restricting their utility to closed-set scenarios~\cite{lee2023languageinformed}. These limitations hinder broad applicability and therefore demand further resolution. Therefore, in this paper, we consider each visual entity to be the composition of the ``object'' concept and multiple ``attributes'' defined along different attribute axes. 

Unlike previous approaches~\cite{chefer2023hidden, vinker2023concept,lee2023languageinformed} that treat object and non-object concepts equally, we reformulate the concept hierarchy by distinguishing between \xzadd{attribute~\cite{farhadi2009describing} and object} concepts. In this hierarchy, a physical entity is primarily represented as an object, characterized by various axes
of attributes. Based on this, we introduce a new task, namely \emph{customized concept decomposition}, to address the limitations of the current work. Specifically, we aim to leverage diffusion models~\cite{ho2020denoising} to disentangle concepts of the ``object'' and the ``attributes'' along different axes. The disentangled concepts are learned within token embeddings, which can then be embedded into prompts to generate concept-specific images. To determine the attribute axes that the model learns, we enable the model to follow human specifications of the axes in the form of natural language, such as ``color" or ``age". 
This task encompasses two \emph{open-world} properties that pose challenges: \textbf{(1)} attribute axes are open, meaning there are no restrictions on the range of axes, allowing for the free specification of attribute axes; and \textbf{(2)} concept vocabulary is open, where object and axis-wise attribute ranges are not predefined, necessitating the automatic learning of attributes present within the image.



To address these problems, we present \model, for \xzadd{\underline{Cus}tomized Visual \underline{Concept} Decomposition with Diffusion Models}.
\model consists of two training stages: vocabulary-guided concept decomposition and joint concept refinement. The two stages enable the extraction of decomposed concepts of the object and attributes into distinct token embeddings. 
In the first stage, our objective is to obtain vocabulary anchors based on user specifications and acquire concept embeddings by learning anchor weights.
Our method first leverages LLMs to obtain a vocabulary corresponding to each specified attribute axis. 
We then automatically retrieve the words that interpret the attribute by learning a linear projection in the textual space through diffusion optimization. 
The decomposed concepts are extracted into token embeddings, referred to as concept centroids, which are obtained by computing a learnable weighted sum of \xzadd{retrieved words from the vocabulary}.
The weights learned in this stage naturally facilitate concept retrieval, explicitly predicting the concept categories of the image, which emerges as a beneficial byproduct of our framework.
In the second stage, we conduct multi-token Textual Inversion~\cite{gal2022image} to jointly fine-tune all concept embeddings obtained through the weighted sum of \xzadd{retrieved words}. This stage expands the semantic information captured by concept embeddings, originally constrained by a simple weighted sum of \xzadd{retrieved words}, thereby enriching the expressiveness of concept generation.

Finally, we devise an evaluation benchmark for properly assessing the customized concept decomposition task. This benchmark incorporates a dataset collected from VAW-CZSL~\cite{saini2022disentangling}, and includes tailored evaluation metrics for evaluating the performance of our model from three aspects: generation fidelity, embedding similarity, and retrieval accuracy. Through extensive qualitative and quantitative experiments, we demonstrate that our model establishes the state-of-the-art performance.

In summary, our contributions are as follows:
    (1) We tackle a new and important task, namely customized concept decomposition, aiming to decompose the object and the attribute along human-specified attribute axes.
    (2) We present \model, the first method to resolve open-world concept decomposition by incorporating two training stages, \ie, vocabulary-guided concept decomposition and joint concept refinement. \model simultaneously facilitates the decomposition, retrieval, and generation of concepts.
    (3) We introduce an evaluation benchmark for assessing the performance of \model in customized concept decomposition, on which \model establishes state-of-the-art performance.

\section{Related Work}

\subsection{Text to Image Generation}
\xzadd{The task of text-driven image generation has been studied in the literature, starting from the GAN-based frameworks~\cite{goodfellow2014generative}. Models utilizing techniques like attention mechanisms~\cite{xu2018attngan} and cross-modal contrastive methods~\cite{zhang2021cross, ye2021improving} achieve good results. Rich visual outcomes text-to-image generation is accomplished through auto-regressive models~\cite{ramesh2021zero, yu2022scaling} trained on large-scale text-image datasets.}

\textbf{Diffusion models}~\cite{ramesh2022hierarchical, nichol2021glide, saharia2022photorealistic, rombach2022high} have achieved extensive attention. In contrast to the conditional model training approach, several methods now leverage test-time optimization to navigate the latent space of an already trained generator~\cite{rombach2022high, crowson2022vqgan}, \xzadd{often using a guide like} CLIP~\cite{radford2021learning}. Recently, Latent Diffusion Models (LDMs)~\cite{rombach2022high} represent an advancement in diffusion technology by operating in a compressed latent space, utilizing a denoising diffusion probabilistic model (DDPM) ~\cite{ho2020denoising}. Building upon the foundation of LDMs, Stable Diffusion Models focus on enhancing the stability and efficiency of the image generation process. 

There are many newly developed large-scale models for converting text to images. Imagen ~\cite{saharia2022photorealistic}, DALL-E2 ~\cite{ramesh2022hierarchical}, Parti ~\cite{yu2022scaling}, CogView2 ~\cite{ding2022cogview2}, Stable Diffusion ~\cite{rombach2022high} and GPT4 ~\cite{achiam2023gpt}, have showcased remarkable capabilities in semantic image generation. However, these models primarily rely on textual prompts and lack the ability to \xzadd{maintain the subject's identity consistently} and offer detailed control over the nuances of the generated images.
\subsection{Textual Inversion}
Recent advances~\cite{liu2023more, avrahami2023blended, nichol2021glide, choi2021ilvr,hertz2022prompt} in generative models have explored various strategies to enhance controllability, enabling the generation of images based on specific subjects and guided by prompts, while preserving the unique identity of the subject. \xzadd{Gal~\etal~}\cite{gal2022image} proposed Textual Inversion, where visual concepts, such as objects or styles, are represented via the introduction of tokens within the embedding space of a static text-to-image model, enabling the creation of compact, personalized token embeddings. DreamBooth~\cite{ruiz2023dreambooth} allows for the integration of the subject into the output range of the model, leading to the creation of new images that maintain the essential visual characteristics of the subject. Custom Diffusion~\cite{kumari2022customdiffusion} addresses \xzadd{learning from a few examples and concept composition}. It refines only a selected portion of the parameters in the cross-attention layers, markedly decreasing the time required for fine-tuning. \xzadd{There are also many works further improve Textual Inversion~\cite{tewel2023key, wei2023elite, Hao2023ViCo, ma2023unified, shi2023instantbooth, jia2023taming, gal2023encoder, chen2024subject}. They enhance the applicability of Textual Inversion by extending it to more text-to-image generation tasks, improving image quality, making the generated images more realistic, and speeding up the process, facilitating future work.}

\subsection{Visual Concept Composition}
\paragraph{\bf Compositional zero-shot learning.}
This task aims to recognize text information from an image, where the text information refers to novel combinations of known attributes and objects
during training. 
Previous works ~\cite{li2020symmetry, mancini2021open, mancini2022learning, misra2017red, naeem2021learning, nagarajan2018attributes, purushwalkam2019task, wei2019adversarial} have attempted to merge the embeddings of attributes and objects, projecting them onto a shared domain that encompasses both word and image representations. Newer studies ~\cite{li2022siamese, hao2023learning, ruis2021independent, saini2022disentangling} have explored the concept of visual disentanglement with notable success. Recently CSP~\cite{nayak2023learning} introduced CLIP ~\cite{radford2021learning} to tackle this task and many works use it as a base model and further enhance the outcome~\cite{zheng2024caila, Lu_2023_CVPR}.

\paragraph{\bf Visual concept decomposition in generative models.}
Many works focus on decomposition along specific axes using generative models. For example,~\cite{lee2023languageinformed} focuses on three axes, color, material, and category. It \xzadd{distills knowledge} from pre-trained vision-language models. After training separate encoders, the concept encoders extract disentangled concept embeddings along various concept axes specified by the language.~\cite{wang2023styleadapter, sohn2023learning} mainly focus on style.~\cite{wang2023styleadapter} introduces a method devoid of LoRA for generating stylized images, which, through the use of textual prompts and style reference images, \xzadd{produces an output image in a
single pass}.~\cite{sohn2023learning} focuses on the challenge of domain-adaptive image synthesis, which involves instructing pre-trained image generation models to adopt a new style or concept with minimal input, sometimes as little as a single image, to create new images.

\section{Method}

\begin{figure*}[tb]
  \centering
  \includegraphics[width=0.95\linewidth]{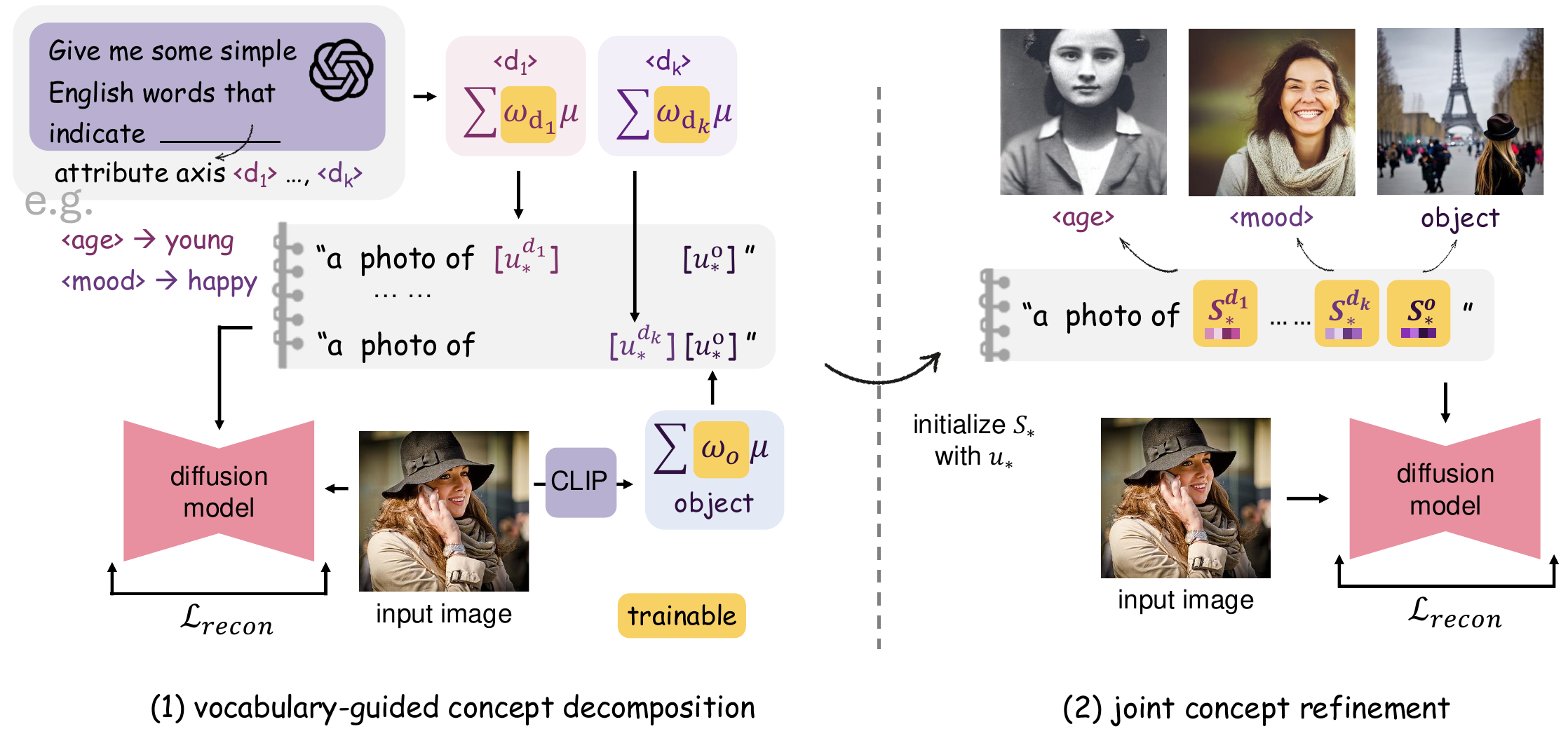}
  \caption{\textbf{Pipeline.} Given an input image and user-specified attribute axes, we aim to decompose the visual concepts including the object concept and the attributes concepts along the specified axes. Our method encompasses two stages.
  \textbf{(1)} To obtain concept vocabularies, we query an LLM (like ChatGPT~\cite{achiam2023gpt}) to derive axis-wise attribute vocabularies, and examine CLIP similarities between object nouns and the input image to derive object vocabularies.  
  On the derived vocabularies, we train learnable \xzadd{anchor} weights, such as $w_{d_k}$ on the $d_k$ \xzadd{attribute} axis and \xzadd{$w_o$} for the object concept, to select and aggregate the corresponding token embeddings.
  \textbf{(2)} With the aggregated token embedding $u_\star$, which represents the concept centroid of the object or its attributes along specific axes, we further fine-tune them jointly to enhance the \xzadd{fidelity} and quality of generation. The fine-tuned token embeddings, such as $S_\star^{d_k}$ and \xzadd{$S_\star^o$}, can be inserted into text prompts for concept generation.}
  \label{fig:pipeline}
\end{figure*}
\hsz{Given an image $x$ that contains an object characterized by attributes from various axes, one can specify one or more attribute axes $\mathbf{D} = \{d_1, \ldots, d_k\}$ regarding the image\xzadd{,~\eg, age in} \cref{fig:pipeline}. Our objective is to decompose the object and attributes along the specified axes within $\mathbf{D}$ and learn distinct token embeddings for each of them. Under our open-world setting, $\mathbf{D}$ is a subset of an open set, within which any attribute axis ${d_i}$ is permissible. To address this problem, we first discover concept anchors by querying an LLM with a prompt of the specified concept axis. We then learn the concept centroid by integrating all discovered anchors. Afterward, we jointly fine-tune the integrated token embeddings to capture richer semantic information from the image. In the following subsections, we introduce our model \model that incorporates two training stages: vocabulary-guided concept decomposition \xzadd{in \cref{sec:step1}} and joint concept refinement \xzadd{in \cref{sec:step2}}.
We present our training objective in \cref{sec:detail}. An overview of our method is shown in \cref{fig:pipeline}.}

\paragraph{\bf Preliminary.}
\hsz{Latent Diffusion Model (LDM)~\cite{rombach2022high} is a latent text-to-image diffusion model derived from Diffusion Denoising Probabilistic Model (DDPM)~\cite{ho2020denoising}, optimized with the denoising loss:} 
\begin{equation}
  \mathcal{L}_{LDM} := \mathbb{E}_{z \sim \mathcal{E}(x),y,\epsilon \sim \mathcal{N}(0,1),t} \left[\left\| \epsilon - \epsilon_\theta (z_t, t, c_\theta (y)) \right\|^2_2\right],
  \label{eq:ldm}
\end{equation}
where \(\mathcal{E}\) is an encoder mapping \xzadd{image} \(x\) into a spatial latent code \(z\),
$z_t$ is the latent noised to time $t$, $\epsilon$ is the unscaled noise sample, $\epsilon_\theta$ is the denoising network, and \( c_{\theta} \) is a model that maps the text prompt $y$ into textual embeddings. $c_{\theta}$ and $\epsilon_\theta$ are jointly optimized \xzadd{during} training.
Based on LDM, Textual Inversion~\cite{gal2022image} aims to capture the characteristics of a single visual concept from a small set of training images (typically 3-5). Textual Inversion learns a token embedding \textit{\(S_*\)} to represent the concept by optimizing the loss defined in \cref{eq:ldm}, while fixing $c_{\theta}$ and $\epsilon_\theta$.
In our task, we aim to learn a token embedding of the object string \( S^o_* \) that represents the decomposed object concept, and multiple attribute token embeddings of the \xzadd{attribute string set} \( \{S^{d_i}_* \ |\ d_i \in \mathbf{D} \} \) that represent decomposed attributes along multiple axes specified by $\mathbf{D}$.


\subsection{Vocabulary-guided Concept Decomposition}
\label{sec:step1}
\xzadd{We aim for the visual concept to be decomposed along the axes user specify,} and we achieve this by two stages: (1) acquiring a knowledge base corresponding to the specified attribute axes through querying an LLM, such as ChatGPT~\cite{achiam2023gpt}, and (2) selecting and aggregating the vocabularies from the knowledge base to obtain the concept centroids by optimizing a learnable weighted-sum mechanism.

\paragraph{Axis-specific knowledge acquisition.}
For each specified attribute axis ${d_i}$, we input the prompt ``\texttt{Give me some simple English words that indicate <${d_i}$>}'' into an LLM, prompting it to provide us with specific knowledge base related to the given axes,~\eg, \xzadd{``\texttt{Give me some simple English words that indicate age}''}. This knowledge base is constructed as a curated set of attribute words \xzadd{along} the axis ${d_i}$. We denote the resulting set of attribute words within the axis $d_i$ as $\mathbf{A}^{d_i} = \{a_j^{d_i}\}_{j=1}^N$, where a total of $N$ attribute words are acquired.
The acquisition covers the majority of attribute vocabularies used to describe an object along the specific axis. We leverage them in the subsequent optimization to guide our model to learn axis-specific features within this vocabulary domain, enabling concept decomposition along the attribute axis.

We also construct a word list for decomposing the object concept. An object can be described by various vocabularies; for instance, ``a house'' may also be referred to as a ``cabin'', ``cottage'', or ``lodge''. Therefore, we consider all words covered in the tokenizer of the CLIP~\cite{radford2021learning} text encoder.
Due to the large scale of the word set, we employ a two-stage filtering process to select a suitable subset of object words: (1) first, we examine CLIP similarity between images and candidate words to pick out words that match the image entity; (2) second, we use \xzadd{NLTK}~\cite{bird2009natural} to select nouns for constructing the final resulting set. We denote the resulting word set as $\mathbf{O} = \{o_p\}_{p=1}^M$, in which we obtain $M$ object words in total.

\paragraph{Concept vocabulary aggregation.}
We aim to aggregate the concept vocabulary along each axis. To achieve this, we train a projection MLP for each axis that maps token \xzadd{embeddings} to a 1-dimensional weight value. \xzadd{Through experiments, we find that directly training the weights of words in the vocabulary fails to capture the target content. In other words, the images generated using the learned embeddings contain almost no information.} But the MLP demonstrates a stronger learning ability than direct weight training. We then obtain a weight for each candidate word and ultimately compute the weighted sum of token embeddings. Specifically, let $\mu(\cdot)$ be the embedding encoder that maps a word to a $\ell$-dimensional token embedding, $\omega_{d_i}$ be a projection MLP for the $d_i$-axis attribute word, and $\omega_{o}$ be a projection MLP for the object word. $\omega_{d_i}$ and $\omega_{o}$ map a $\ell$-dimensional token embedding to a 1-dimensional weight value, denoted as $w_j^{d_i}$ and $w_p^o$ respectively. 
We can then compute the weighted sum of top token embeddings. We
define ``top'' by ranking all tokens \xzadd{along the same axis} by weight. Considering only those with the high weights, we perform a weighted sum on their embeddings. The number of tokens we consider is $N'$ for the attribute and $M'$ for the object. Therefore, the computation of weighted sum is given by
\begin{align}
 u_*^{d_i} &= \sum_{j=1}^{N'} \textcolor{BrickRed}{\underbrace{\omega_{d_i}(\mu(a_j^{d_i}))}_{w_j^{d_i}}} \cdot \mu(a_j^{d_i}), \ d_i \in \mathbf{D} \\
 u_*^{o} &= \sum_{p=1}^{M'} \textcolor{NavyBlue}{\underbrace{\omega_{o}(\mu(o_p))}_{w_p^o}} \cdot \mu(o_p)
\end{align}
where $\{u_*^{o},u_*^{d_i}\ |\ d_i \in \mathbf{D}\}$ is the derived set of embedding vectors that respectively represent the $d_i$-axis attribute and the object. It is worth noting that the word that corresponds to the highest weight can be regarded as an explicit prediction of the concept category along the specific axis, which
emerges as a beneficial byproduct. \xz{Each weight indicates the likelihood that the corresponding word describes the image, and the weighted sum provides a \xzadd{combination} of multiple words within that dimension.}
\subsection{Joint Concept Refinement}
\label{sec:step2}
\hsz{The embedding vectors within the set $\{u_*^{o},u_*^{d_i}\ |\ d_i \in \mathbf{D}\}$ can directly serve as representations of our decomposed concepts. However, these embeddings are derived from a simple weighted sum of token embeddings from existing vocabularies, limiting their capacity to capture and convey the complete characteristics of the unique entity in the given image. To address this issue, we use the token embeddings obtained through the weighted sum to initialize new learnable token embeddings, and then directly optimize these token embeddings by \xzadd{jointly} training.}

Specifically, let the concept strings in \( \{S^{o}_*, S^{d_i}_* \ |\ d_i \in \mathbf{D} \} \) initially correspond to values of the token embeddings in $\{u_*^{o},u_*^{d_i}\ |\ d_i \in \mathbf{D}\}$. We then optimize these token embeddings by conditioning the denoising process on the text prompt ``\texttt{a photo of $S_*^{d_1}$ $S_*^{d_2}$ $\ldots$ $S_*^{d_k}$ $S_*^{o}$}''. In this training process, all token embeddings corresponding to the concept strings are jointly refined to better capture the complete characteristics from the image. After joint concept refinement, we obtain token embeddings that accurately represent the object concept and the attribute concepts along specific axes, serving as the final outcome of the customized concept decomposition.

\subsection{Training Objective}
\label{sec:detail}
In the stage of vocabulary-guided concept decomposition, we optimize the projection MLPs by minimizing:
\begin{equation}
  \mathcal{L}_{\text{stage1}} := \mathbb{E} \left[\left\| \epsilon - \epsilon_\theta (z_t, t, c_{\textcolor{PineGreen}{{\{\omega_{o}, \omega_{d_i} | d_i \in \mathbf{D}\}}}} (y)) \right\|^2_2\right],
  \label{eq:loss1}
\end{equation}
where \textcolor{PineGreen}{${\{\omega_{o}, \omega_{d_i} | d_i \in \mathbf{D}\}}$} are learnable. In the stage of joint concept refinement, after being initialized with $\{u_*^{o},u_*^{d_i}\ |\ d_i \in \mathbf{D}\}$, the token embeddings corresponding to the concept strings in \( \{S^{o}_*, S^{d_i}_* \ |\ d_i \in \mathbf{D} \} \) are optimized using
\begin{equation}
  \mathcal{L}_{\text{stage2}} := \mathbb{E} \left[\left\| \epsilon - \epsilon_\theta (z_t, t, c_{\textcolor{RawSienna}{\{S^{o}_*, S^{d_i}_*| d_i \in \mathbf{D} \}}} (y)) \right\|^2_2\right],
  \label{eq:loss2}
\end{equation}
where \textcolor{RawSienna}{$\{S^{o}_*, S^{d_i}_*| d_i \in \mathbf{D} \}$} are learnable and finally derive the final generative representations of the decomposed object and attribute concepts.

\section{Experiments}

\subsection{Dataset and Evaluation}
\paragraph{\bf Dataset.} 
In our task, we need to distinguish between the \xzadd{object} and attributes in images, which has been studied in the task of CZSL~\cite{misra2017red, nagarajan2018attributes, purushwalkam2019task, wei2019adversarial}. Therefore, we choose the high-quality CZSL dataset VAW-CZSL~\cite{saini2022disentangling} for \xzadd{our experiments}. Specifically, we collect 56 images from VAW-CZSL, labeling them with the ground truth, \eg, ``young woman''. The labels include 14 attributes and 33 objects, with the attributes categorized into 8 attribute axes (\eg, ``age''). Due to the fact that CZSL only considers a single attribute, in our \xzadd{quantitative} experiments we only take into account a single attribute + object combination. We also collect two prompt templates, one for attributes and another for objects, comprising 11 prompts respectively. Details are provided in the supplementary material. During training, we apply our method on each image, \xz{without the need for a large training dataset.}

\paragraph{Evaluation metrics.} We assess concept decomposition from three perspectives: visual fidelity, textual alignment, and retrieval accuracy. Specifically, we apply three evaluation metrics: CLIP-I, SIM\textsuperscript{emb}, and ACC.
\begin{itemize}[topsep=5pt,partopsep=0pt, parsep=0pt, left=0cm]
\item \textbf{CLIP-I} computes the similarity between the CLIP~\cite{radford2021learning} features of the input image and \xzadd{the generated images}, assessing how well the generated images retain subject details. We report CLIP-I by examining the images generated by three prompts: one with only the object (o), one with only the attribute (a), and one with both (a+o). CLIP-I is evaluated under the prompt with both the attribute and object is the core metric in our task.
\item \textbf{SIM\textsuperscript{emb}} computes the CLIP~\cite{radford2021learning} similarity between two \xz{sets of prompts}: one with the attribute (a) or object (o) label of the input image, and another with the aggregated token embedding of the top 5 words after the first stage. This metric evaluates how well the aggregation weights learned on the vocabulary aligns with the \xzadd{ground-truth} text.
\item \textbf{ACC} assesses the retrieval accuracy of our learned concept embeddings. Specifically, we compute the CLIP similarity between two prompts: one with the attribute (a) or object (o) words, and another with the trained embeddings after the ``concept vocabulary aggregation'' stage. 
We consider all 440 attributes and 541 objects in VAW-CZSL~\cite{saini2022disentangling}. For each input image, we retrieve the word with the highest similarity, check if it matches the label of the input image, and then \xzadd{measure the retrieval performance}.
\end{itemize}


\subsection{Implementation Details}
\begin{figure}[tb] 
  \centering
  \includegraphics[width=\linewidth]{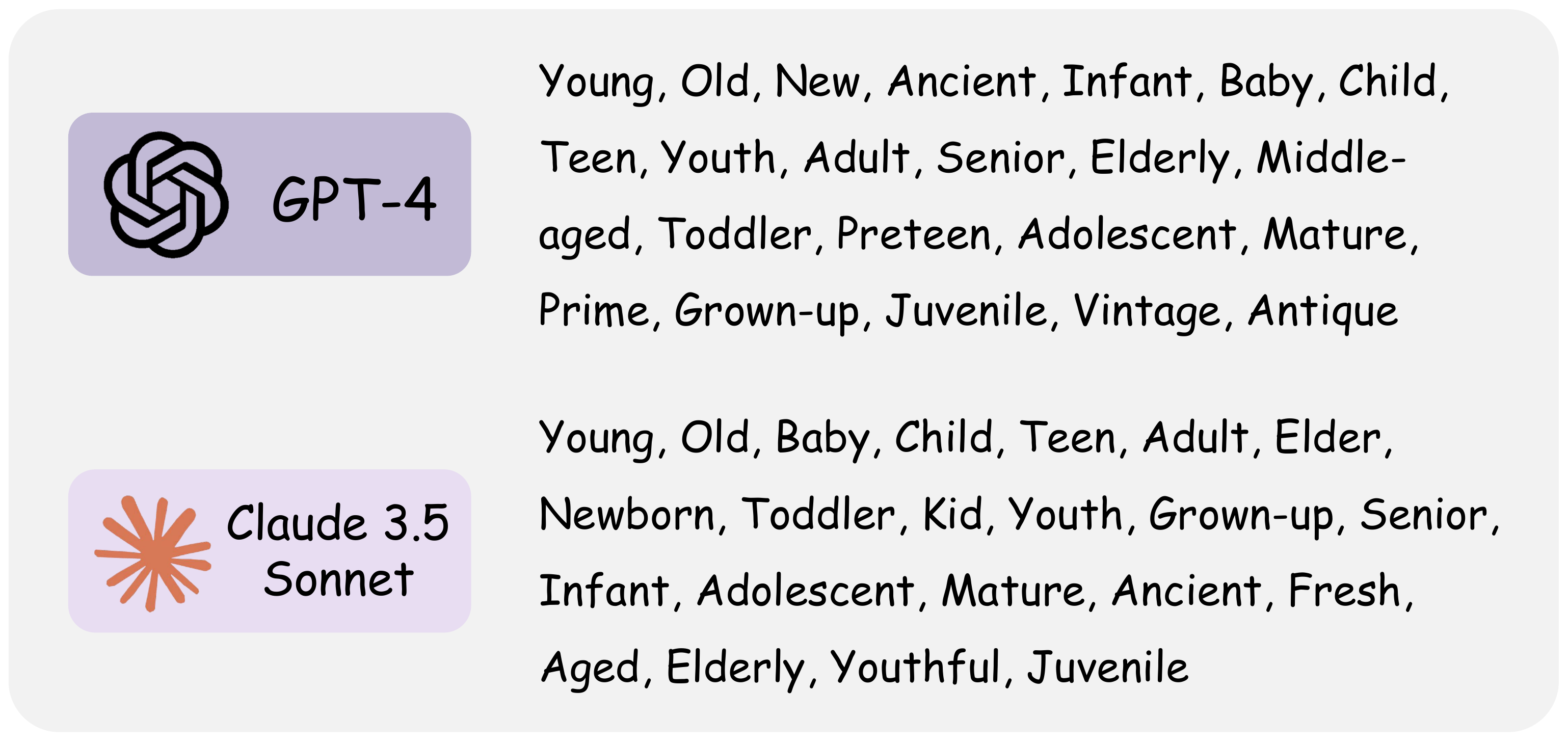}
  \caption{\textbf{Comparison between LLMs.} Taking the \textit{age} attribute axis as an example, we compare the words generated by GPT-4 and Claude 3.5 Sonnet.} 
  \label{fig:gpt}
\end{figure}

\paragraph{Implementation.}
\label{sec:implement}
\hsz{In this paper, We use Stable Diffusion v2-1~\cite{rombach2022high} as our base model and use the ViT-Base-32~\cite{dosovitskiy2021imageworth16x16words} backbone for the CLIP model to extract image features and filter the words of the object. We \xzadd{adopt a} 4-layer MLP for weight projection. Regarding the size of the vocabulary, $N$ is set to 22 for attribute vocabulary, and $M$ is set to 500 for object vocabulary. In weighted sum, we only consider the top $N'$=10 words for the attribute concept and the top $M'$=50 words for the object concept. We use the AdamW optimizer~\cite{loshchilov2017decoupled} to train the model in both stages. In the stage of vocabulary-guided concept decomposition, we use a learning rate of 0.01 for training the attribute projection MLP and a learning rate of 0.001 for the object projection MLP. In the stage of joint concept refinement, we use a learning rate of 0.001 for both attribute and object embeddings. The training of our model takes about 3.5 minutes on a single NVIDIA RTX 3090 GPU.}

\xz{
There are many LLMs available for use, and in this paper, we choose GPT-4. We present a comparison between GPT-4 and \xzadd{Claude 3.5 Sonnet}, showing relatively similar results in~\cref{fig:gpt}. For certain attribute dimensions, the current prompt may sometimes struggle to generate suitable words, but further prompt refinement can address this. As LLMs continue to evolve, we believe they will produce increasingly satisfactory results.
}


\begin{figure}[tb] 
  \centering
  \includegraphics[width=\linewidth]{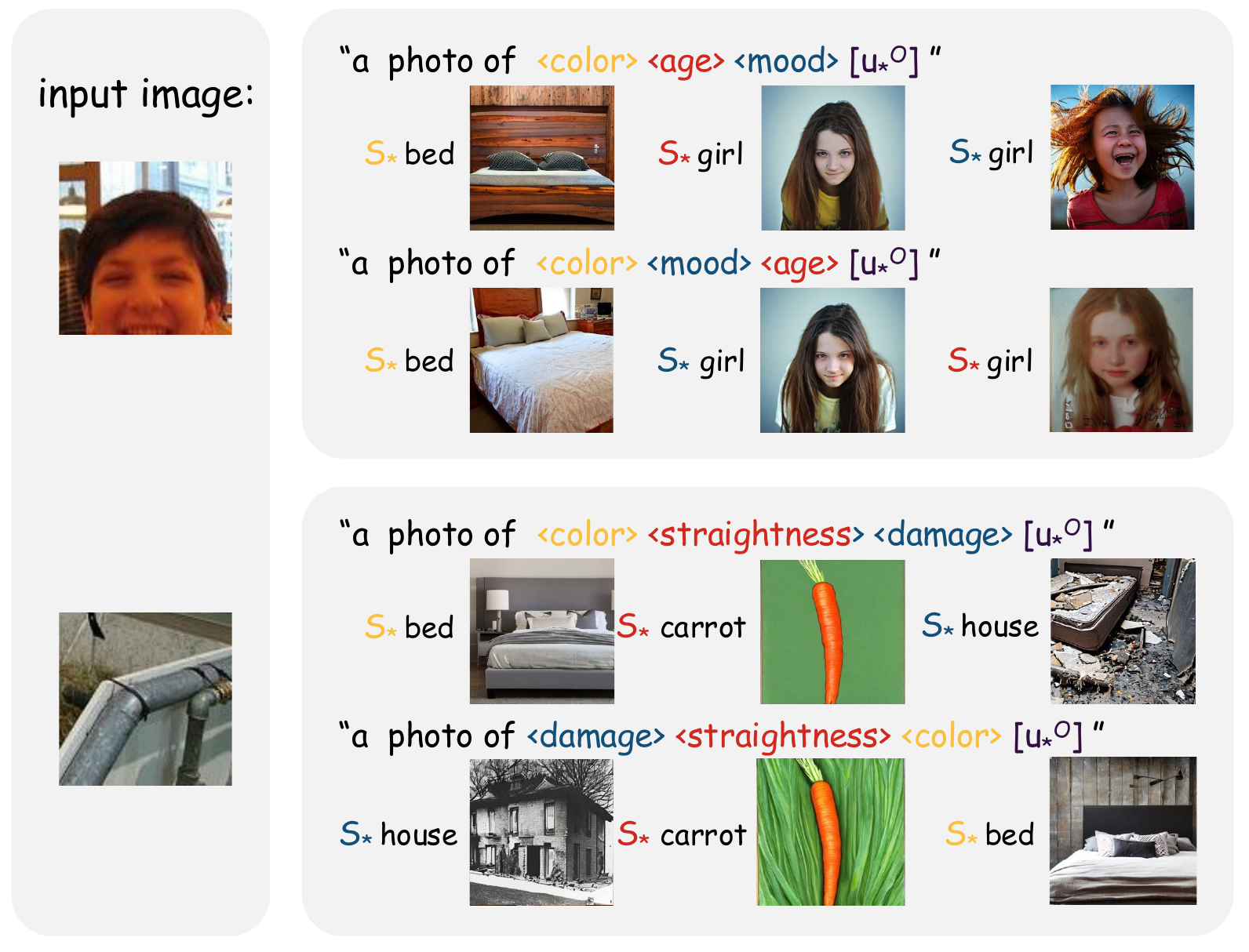}
  \caption{\textbf{Comparison across different orders of attributes.} We present images generated with different orders of attribute axes.}
  \label{fig:order}
\end{figure}

\xz{In this method, we do not explicitly define the order of the attribute axes. Our qualitative experiments in~\cref{fig:order} demonstrate that the order does not significantly impact the quality of the generated images. This is because, technically, the cross-attention modules in Stable Diffusion process all words equally.}

\subsection{Comparison}

\begin{table}[tb]
  \caption{Quantitative comparison among TI\textsuperscript{$\alpha$}, TI\textsuperscript{$\beta$} and our method. TI\textsuperscript{$\alpha$} uses the ground-truth labels for training, representing the loose upper limit of performance. `a+o' represents generating images use both attribute and object prompts, while `a' and `o' use only attribute or object prompts}
  \label{tab:1}
  \centering
  \setlength{\tabcolsep}{5pt}
  \scalebox{0.85}{
  \begin{tabular}{lccccccc}
    \toprule
      Method & \multicolumn{3}{c}{CLIP-I} & \multicolumn{2}{c}{SIM\textsuperscript{emb}} & \multicolumn{2}{c}{ACC}\\
      \cmidrule(lr){2-4} \cmidrule(lr){5-6} \cmidrule(lr){7-8}
      & a+o & a & o & a & o & a & o \\
      \midrule    
      \textcolor{gray}{TI\textsuperscript{$\alpha$}} & \textcolor{gray}{0.701} & \textcolor{gray}{0.518} & \textcolor{gray}{0.593} & \textcolor{gray}{0.836} & \textcolor{gray}{0.746} & \textcolor{gray}{0.964} & \textcolor{gray}{0.661}\\
      \hdashline 
      {TI\textsuperscript{$\beta$}} & 0.458 & 0.484 & 0.492 & 0.642 & 0.587 & 0.000 & 0.000 \\
      Ours & \textbf{0.701} & \textbf{0.511} & \textbf{0.641} & \textbf{0.779} & \textbf{0.685} & \textbf{0.411} & \textbf{0.214}\\
  \bottomrule
  \end{tabular}}
\end{table}
\begin{figure}[tb]
  \centering
  \includegraphics[width=\linewidth]{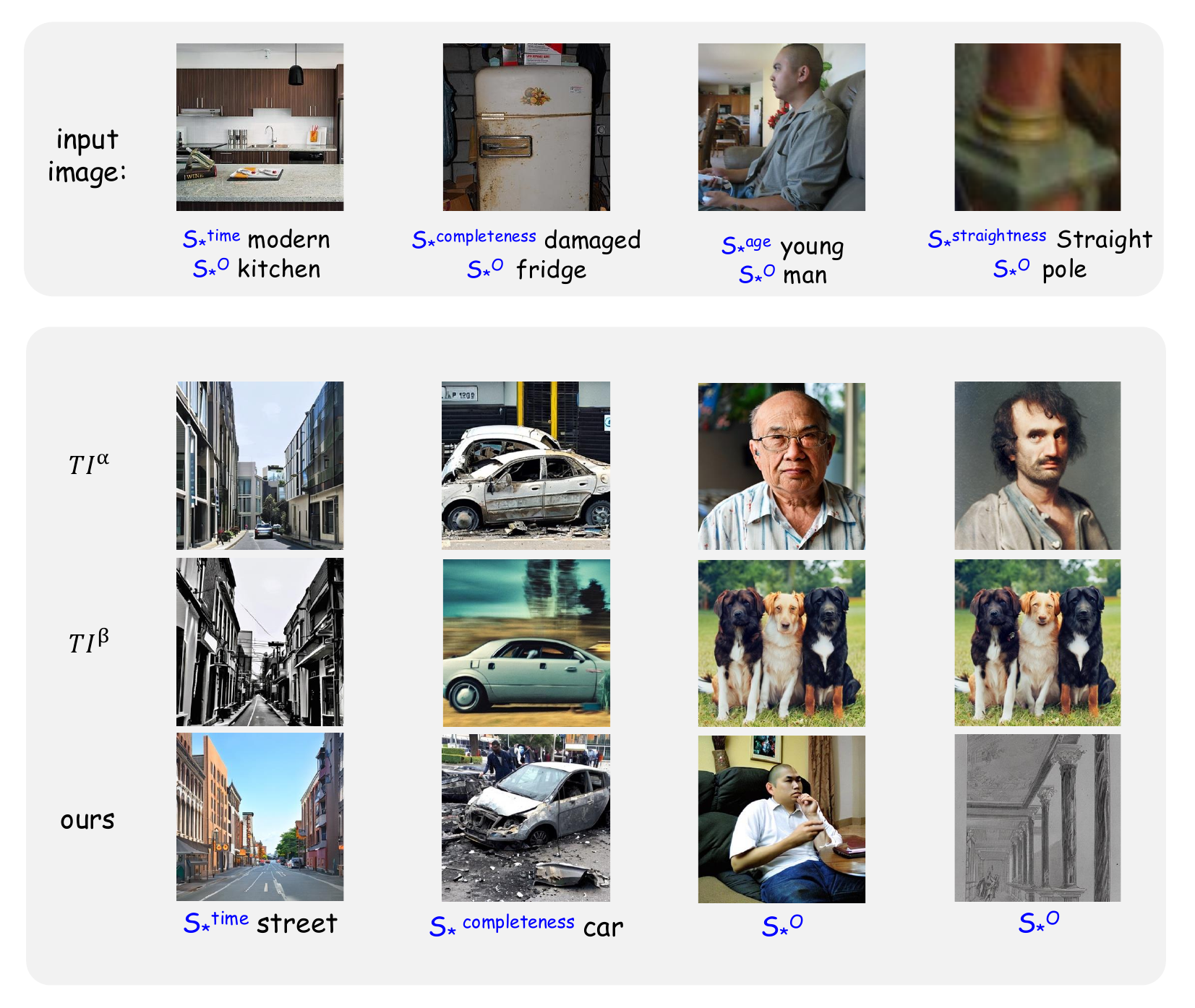}
  \caption{\textbf{Qualitative comparison.} Given one input image (top row), we compare TI\textsuperscript{$\alpha$} (1st row), TI\textsuperscript{$\beta$} (2nd row), and our method (bottom row). We provide the ground-truth labels for the object and \xzadd{attribute} concepts for reference but note that they are not available \xzadd{along} training in \xzadd{TI\textsuperscript{$\beta$} and ours}. Resulting images are generated with the prompt ``\texttt{a photo of}'' followed by the text below the generated image.}
  \label{fig:compare}
\end{figure}

\paragraph{Baselines.} We use Textual Inversion~\cite{gal2022image} as our baseline model for comparison. Since Textual Inversion considers only one visual concept, we adapt it to accommodate multiple concepts. We compare two variants of Textual Inversion. \textbf{(1)} For the first variant, denoted as TI\textsuperscript{$\alpha$}, we insert two placeholders, \(S_*^{1}\) and \(S_*^{2}\), and initialize them with attribute and object labels of the input image, drawing from the ground truth of VAW-CZSL~\cite{saini2022disentangling}. Due to the initialization of ground-truth labels, TI\textsuperscript{$\alpha$} represents the loose upper limit of performance. \textbf{(2)} For the second variant, denoted as TI\textsuperscript{$\beta$}, we directly optimize \(S_*^{1}\) and \(S_*^{2}\) without initialization. This approach is similar to~\cite{vinker2023concept}, but without manual selection.

\paragraph{Quantitative comparison.} We quantitatively compare our model to baselines in \cref{tab:1}. The results indicate our method significantly outperforms TI\textsuperscript{$\beta$}. Compared to TI\textsuperscript{$\alpha$} which \xzadd{has access} to ground-truth labels, our method achieves very close performance in terms of CLIP-I and SIM\textsuperscript{emb} and even performs better on the object concept under CLIP-I. Our method falls short on retrieval accuracy while TI\textsuperscript{$\alpha$} reaches higher because TI\textsuperscript{$\alpha$} initializes learnable tokens with the ground-truth concept words.

\paragraph{Qualitative comparison.} We also present the generated images of our model and baselines in \cref{fig:compare}. We can observe that only our method produce visually reasonable and conceptually aligned images. 
In the last column, TI\textsuperscript{$\alpha$} fails to generate a pole and create a human face. This is because TI\textsuperscript{$\alpha$} initializes the learnable token with the word ``pole''. However, due to its polysemous nature, this ultimately results in the generation of an image depicting a Polish person.
TI\textsuperscript{$\beta$} poorly captures concept information, particularly evident in the last two columns where the generated images closely resemble the initial randomized images, indicating insufficient learning of the second token. We speculate that this is because training both tokens simultaneously without knowledge guidance may mislead the training direction, causing the first token to learn a specific concept embedding while the second token learns a common latent embedding. Overall, the above experimental results demonstrate the effectiveness of our method in utilizing \xzadd{specified axes} and vocabulary knowledge to guide concept decomposition, resulting in superior performance.

\subsection{Ablation Study}
\begin{table}[tb]
  \caption{Quantitative results of ablating joint concept refinement and attribute axes. ``w/o JCR.'' denotes not using joint concept refinement and ``w/o AA.'' denotes discarding attribute axes.}
  \label{tab:ablation}
  \centering
  \setlength{\tabcolsep}{5pt}
  \scalebox{0.85}{
    \begin{tabular}{@{}lccccccc@{}}  
    \toprule
     & \multicolumn{3}{c}{CLIP-I} & \multicolumn{2}{c}{SIM\textsuperscript{emb}} & \multicolumn{2}{c}{ACC} \\
      \cmidrule(lr){2-4} \cmidrule(lr){5-6} \cmidrule(lr){7-8}
    Method & a+o & a & o & a & o & a & o \\
    \midrule
    w/o JCR. & 0.662 & 0.515 & 0.609 & 0.776 & 0.683 & \bf0.446 & 0.196\\
    w/o AA.  & 0.653 & \textbf{0.518} & 0.604& 0.635 & 0.674 & 0.339 & 0.179\\
    ours & \textbf{0.701} & 0.511 & \textbf{0.641} & \textbf{0.779} & \textbf{0.685} & 0.411 & \textbf{0.214}\\
    \bottomrule
    \end{tabular}}
\end{table}

We perform two ablation studies. The first one is that we omit \xzadd{``joint concept refinement'' (\cref{sec:step2})}, focusing solely on \xzadd{``vocabulary-guided concept decomposition'' (\cref{sec:step1})}. The second one is that we do not provide specific attribute axes. Similar to the construction of object vocabulary, the attribute vocabulary is constructed by aggregating words from CLIP~\cite{radford2021learning} tokenizer, following preliminary filtering based on the input image and subsequent adjective extraction using \xzadd{NLTK}~\cite{bird2009natural}.

The quantitative comparisons are presented in~\cref{tab:ablation}. Our full method overall achieves the best performance, especially on both attribute and object concepts (a+o) under the CLIP-I metric. 
This is because adding joint concept refinement can better preserve the complete characteristics of the image, and specifying attribute axes can improve the retrieval of reliable attribute words. 
The setting of ``w/o attribute axes (AA.)'' achieves higher under \xzadd{only attribute (a)} in CLIP-I metric because more word candidates can expand the embedding learning space. This simultaneously leads to a decrease in precision, shown in relatively lower \(SIM^{emb}\) and \(ACC\). The setting of ``w/o joint concept refinement (JCR.)'' achieves higher attribute \(ACC\) because the weighted sum of multiple attribute word embeddings tends to distribute more close to attribute words in natural language.
\begin{figure}[tb] 
  \centering
  \includegraphics[width=\linewidth]{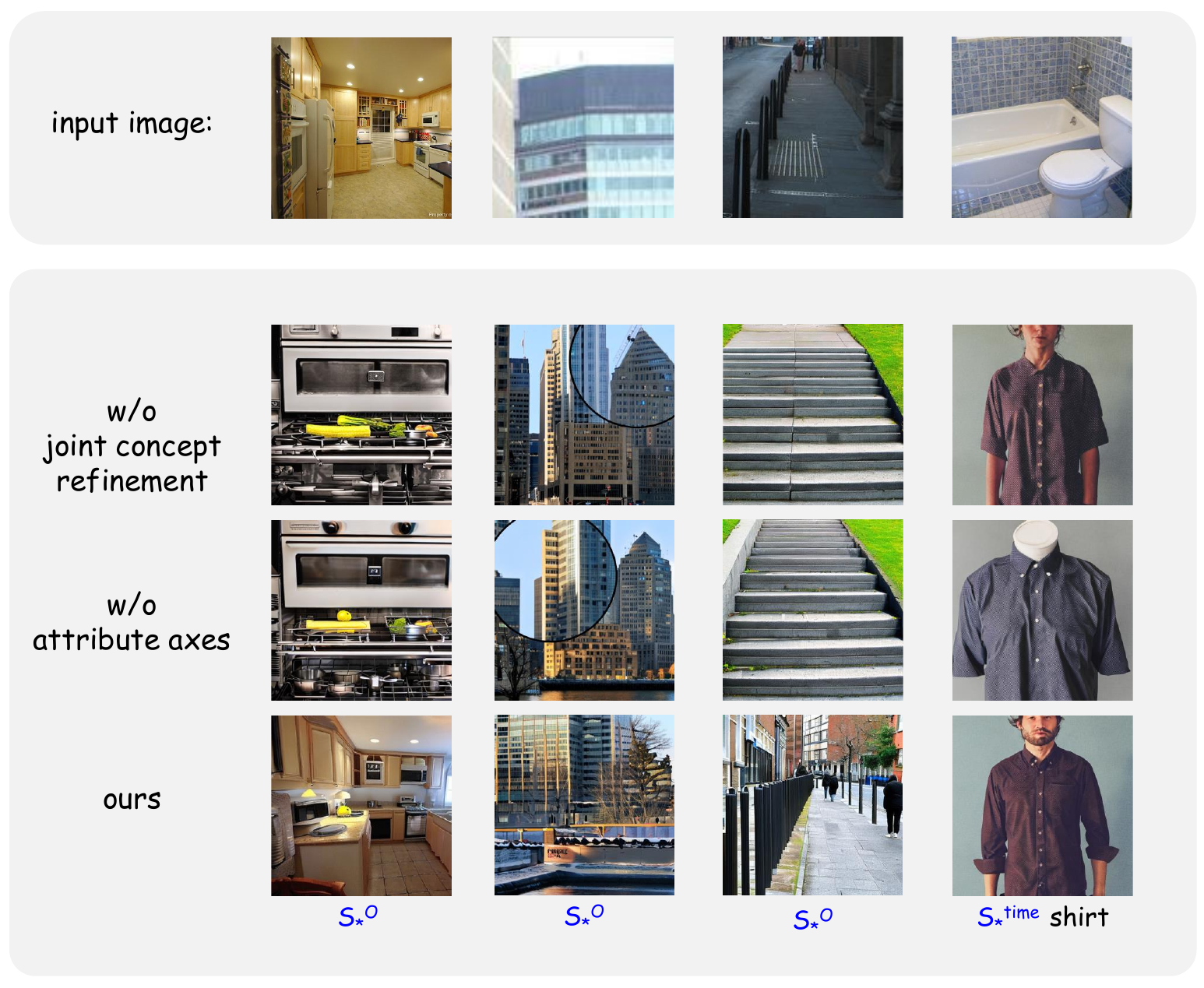}
  \caption{\textbf{Ablation study.} \xzadd{We present a comparison between images generated using object embedding and attribute embedding.} Our method better inherits the visual information of images, rather than generating images based on the corresponding text.} 
  \label{fig:ablation}
\end{figure}

We also show our qualitative results in \cref{fig:ablation}.
In the second column where the input image is an office building, we use visual magnification to highlight that the images generated by the first two methods do not match reality. In contrast, our method generates reasonable and realistic images.

\subsection{Application}

\begin{figure}[tb]
  \centering
  \includegraphics[height=5cm]{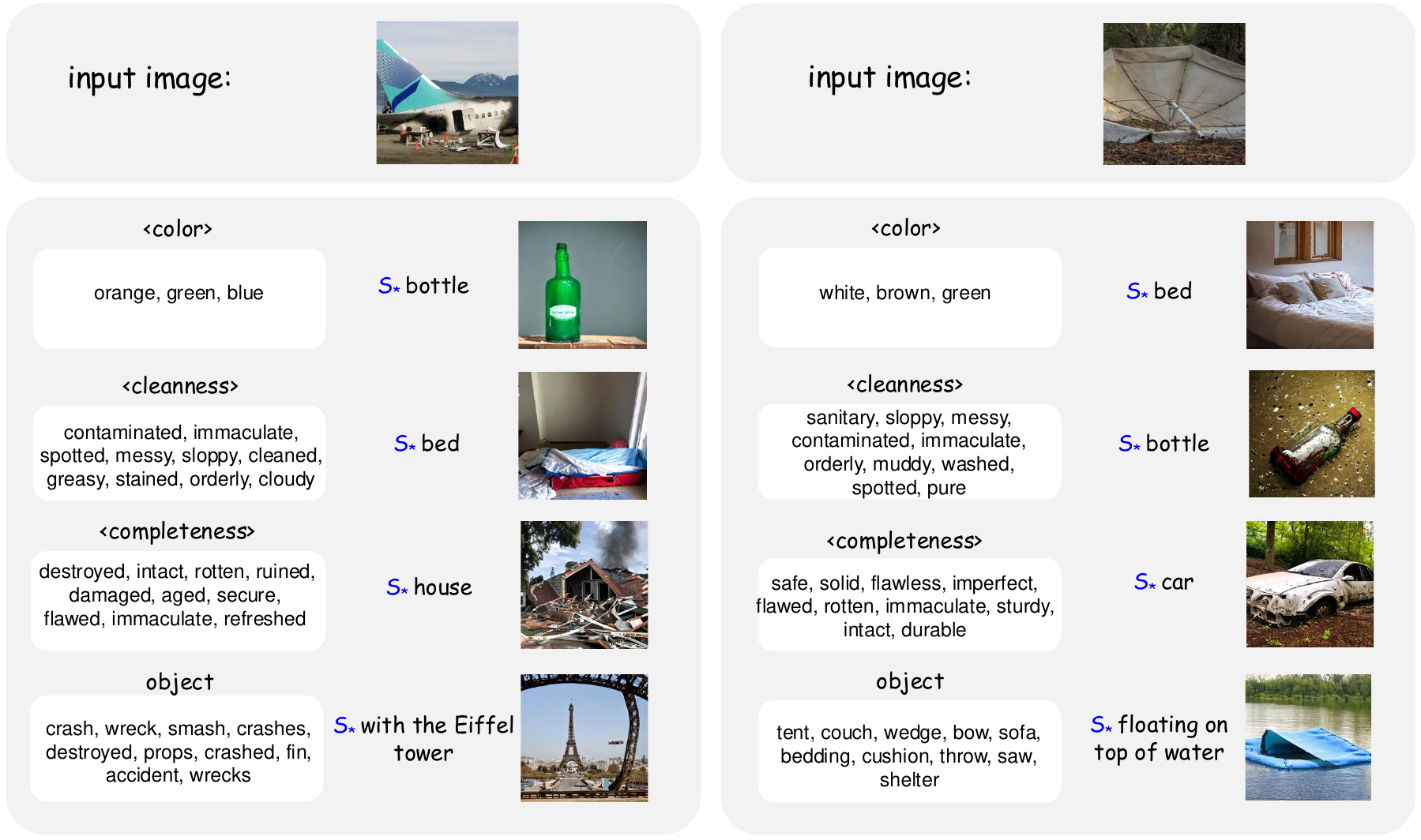}
  \caption{\textbf{Vocabulary prediction and generation.} Concept vocabulary predictions along 3 attribute axes and for the object. We present the top words predicted based on the learned weights (left), and one image generated by each token \textcolor{blue}{\( S_* \)} (right).}
  \label{fig:pic3}
\end{figure}  

\begin{figure}[tb]
  \centering
  \includegraphics[height=4.5cm]{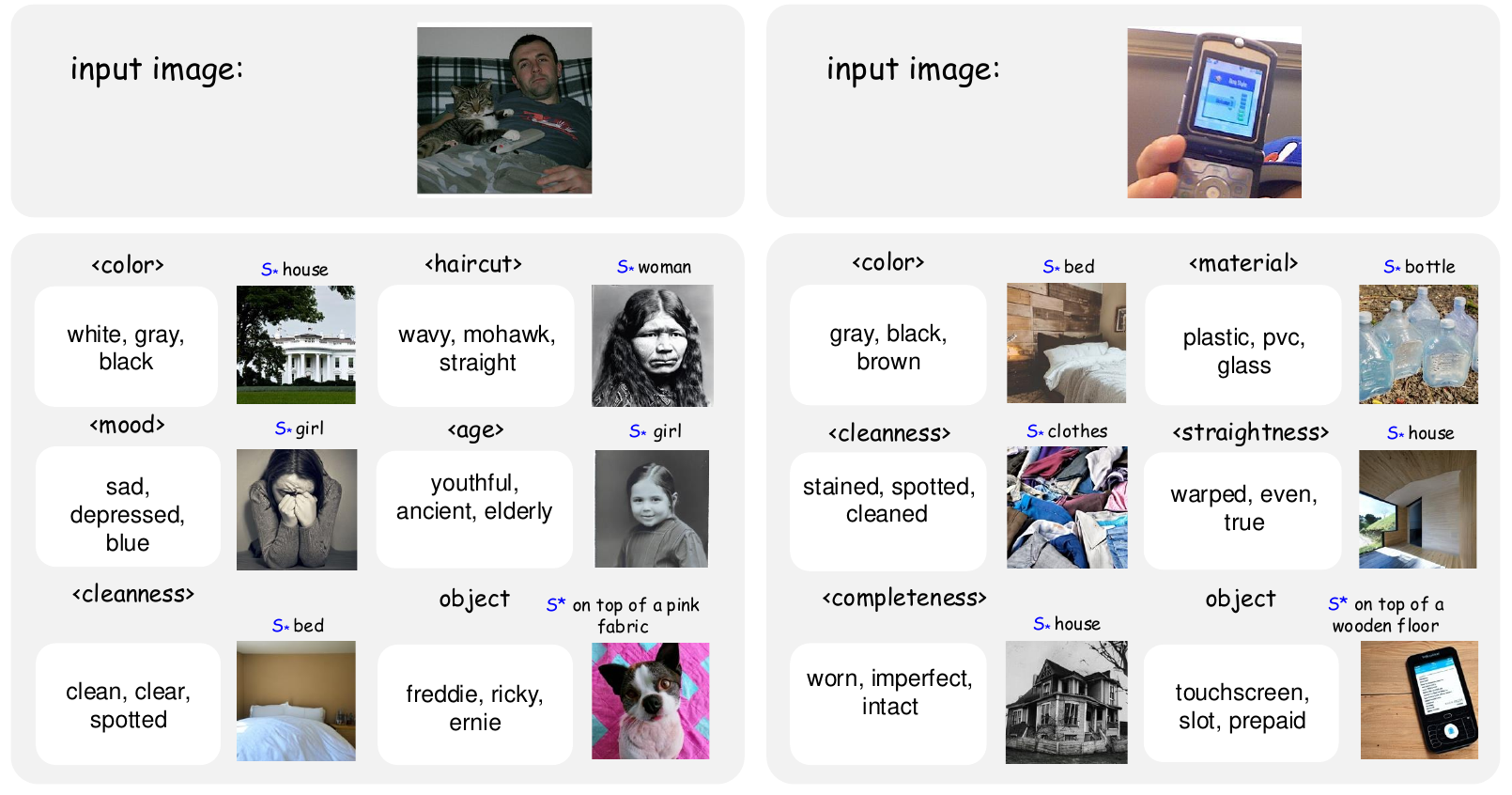}
  \caption{\textbf{Vocabulary prediction and generation.} Concept vocabulary predictions along 5 attribute axes and for the object.}
  \label{fig:pic5}
\end{figure}

\paragraph{Concept vocabulary prediction.} In our training framework, the first stage automatically learns weights for concept word candidates in the vocabulary. Therefore, our model can naturally predict the concept vocabularies of an image once the first-stage training is completed. We present the vocabulary predictions along 3 attribute axes in \cref{fig:pic3} and along 5 attribute axes in \cref{fig:pic5}.
In \cref{fig:pic3}, we specify 3 attribute axes to each input image and learn 3 attribute tokens \( S^{d_1}_* \), ..., \( S^{d_3}_* \), and 1 object token \( S^{o}_* \) during optimization. 
The results indicate that our approach can effectively retrieve the desired words and perform decomposed concept image generation according to user-specified attribute axes. 
In \cref{fig:pic5}, we assign 5 attribute axes to each input image and learn four new tokens \( S^{d_1}_* \), ..., \( S^{d_5}_* \), \( S^{o}_* \) during optimization. \xzadd{The results, compared to \cref{fig:pic3}, show that the words retrieved are less accurate. This suggests that fewer attribute axes can enhance the accuracy of decomposition, making each decomposed concept more complete.}

\paragraph{\bf Concept Removal and Recomposition.} 
\label{sec:remove}
\begin{figure}[tb]
  \centering
  \includegraphics[width=\linewidth]{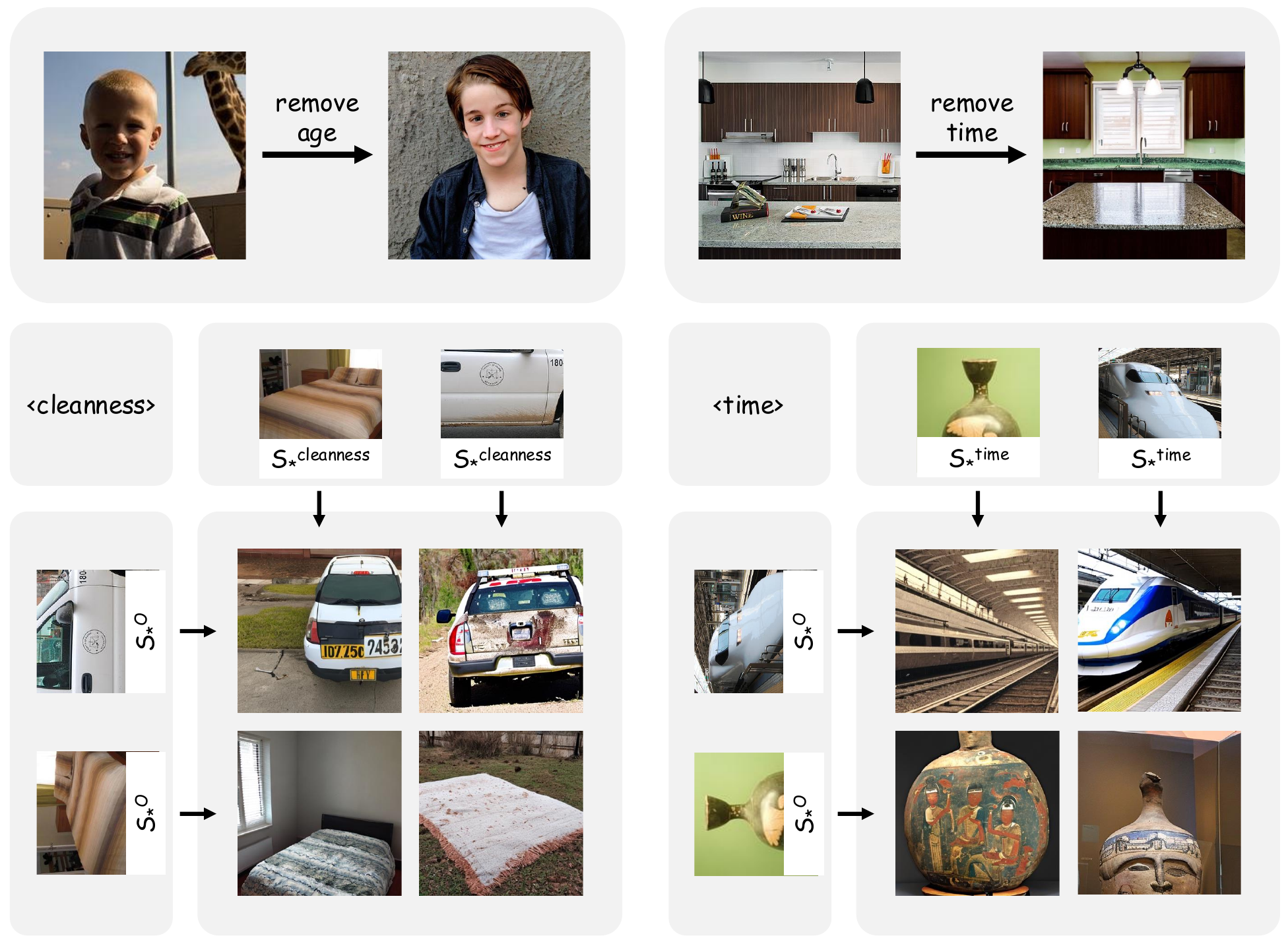}
  \caption{\textbf{\xzadd{Concept removal and recomposition.}} \xzadd{Our method} can remove concepts and generate interesting object without the specified attribute. We are also able to generate novel and meaningful images using the concept from different images.}
  \label{fig:candr}
\end{figure}

\begin{figure}[tb] 
  \centering
  \includegraphics[width=\linewidth]{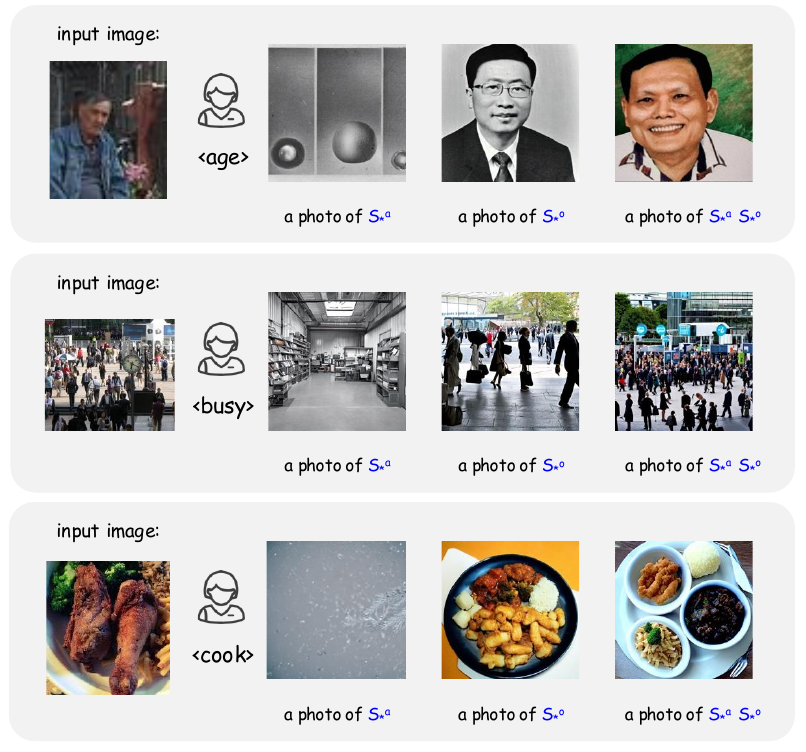}
  \caption{\textbf{Concept generation.} Comparison of generated images of ``attribute'', ``object'' and ``attribute + object''.} 
  \label{fig:ablation2}
\end{figure}

Our method uses user-specified attribute axes as guidance to decompose the input images. Concept characteristics along these attribute axes are learned in specific tokens, with the remaining information captured in the object token.
Thus, by directly using the token embedding of the object to generate images, we can obtain the sole object in the input image to achieve concept removal.
As shown in \cref{fig:candr}, our method can successfully remove attribute concepts and recompose attribute and object concepts from different images, \xzadd{like modern ceramics are the ceramics in the modern museum}. We also show independently generated images with attribute concepts, as well as images combining the attribute and object, in \cref{fig:ablation2}. We observe that attribute concepts alone cannot generate meaningful images, while combining them with object concepts can. This demonstrates that our model effectively isolates object information from attribute tokens. 

Our model also enables the generation of novel images by \xzadd{recomposing} concepts from different images. Specifically, we first learn embeddings of various concepts, which are then combined into prompts to generate novel images that mix these concepts. As shown in \cref{fig:candr}, our method successfully \xzadd{recomposes} visual concepts along specified axes and generates high-quality and meaningful images.

\paragraph{Limitations.}
There is still significant room to improve the consistency of the generated images. 
For instance, in the concept removal case discussed in~\cref{sec:remove}, there appears to be a certain inconsistency between the objects in the generated image and the original one is crucial. Particularly, in the examples of~\cref{fig:candr}, it is not easy to tell whether they are the same boy (top left example) or the same kitchen (top right example). 

\section{Conclusion}
This paper addresses a new and challenging task, namely customized concept decomposition. Given an image and one or more user-specified attribute axes, our goal is to learn token embeddings that represent the object and the attribute along each axis. 
To address this task,  we propose \model, a two-stage framework consisting of vocabulary-guided concept decomposition and joint concept refinement for customized concept discovery. We also curate a dataset and design three evaluation metrics for this task. Through extensive qualitative and quantitative assessments, we demonstrate that our model can effectively address the problem of customized concept decomposition.

\paragraph{Acknowledgment.} This work is supported by the Hong Kong Research Grants Council - General Research Fund (Grant No.: 17211024).

{\small
\bibliographystyle{ieee_fullname}
\bibliography{egbib}
}
\end{document}